\documentclass[journal]{IEEEtran}
\ifCLASSINFOpdf
\else
\fi
\usepackage{cite}
\usepackage{amssymb}
\usepackage{bm}
\usepackage[colorlinks,linkcolor=blue]{hyperref}
\usepackage{graphicx}
\usepackage{float} 
\usepackage{subfigure}
\usepackage{amsmath} 
\usepackage{multirow}
\usepackage{booktabs}

\begin{document}
%
\title{Efficient 3D Point Cloud Feature Learning for Large-Scale Place Recognition}
%
%
%


\author{Le Hui, Mingmei Cheng, Jin Xie, and Jian Yang~\IEEEmembership{Member,~IEEE}
	\thanks{Le Hui, Mingmei Cheng, Jin Xie, and Jian Yang are with the School of Computer Science and Engineering, Nanjing University of Science and Technology, Nanjing 210094, China (e-mail: \{le.hui, chengmm, csjxie, csjyang\}@njust.edu.cn).}
}

%
%

\markboth{Journal of \LaTeX\ Class Files,~Vol.~14, No.~8, August~2015}%
{Shell \MakeLowercase{\textit{et al.}}: Bare Demo of IEEEtran.cls for IEEE Journals}
%



\maketitle

\begin{abstract}
Point cloud based retrieval for place recognition is still a challenging problem due to drastic appearance and illumination changes of scenes in changing environments. Existing deep learning based global descriptors for the retrieval task usually consume a large amount of computation resources ($e.g.$, memory), which may not be suitable for the cases of limited hardware resources. In this paper, we develop an efficient point cloud learning network (EPC-Net) to form a global descriptor for visual place recognition, which can obtain good performance and reduce computation memory and inference time. First, we propose a lightweight but effective neural network module, called ProxyConv, to aggregate the local geometric features of point clouds. We leverage the spatial adjacent matrix and proxy points to simplify the original edge convolution for lower memory consumption. Then, we design a lightweight grouped VLAD network (G-VLAD) to form global descriptors for retrieval. Compared with the original VLAD network, we propose a grouped fully connected (GFC) layer to decompose the high-dimensional vectors into a group of low-dimensional vectors, which can reduce the number of parameters of the network and maintain the discrimination of the feature vector. Finally, to further reduce the inference time, we develop a simple version of EPC-Net, called EPC-Net-L, which consists of two ProxyConv modules and one max pooling layer to aggregate global descriptors. By distilling the knowledge from EPC-Net, EPC-Net-L can obtain discriminative global descriptors for retrieval. Extensive experiments on the Oxford dataset and three in-house datasets demonstrate that our proposed method can achieve state-of-the-art performance with lower parameters, FLOPs, and runtime per frame. Our code is available at \emph{\url{https://github.com/fpthink/EPC-Net}}.
\end{abstract}


%
\IEEEpeerreviewmaketitle

\section{Introduction}
\IEEEPARstart{V}{isual} localization is one of the important tasks in computer vision, which can be applied to a variety of robotic applications such as visual simultaneous localization and mapping (SLAM)~\cite{durrant2006simultaneous,bailey2006simultaneous,cadena2016past}, loop-closure detection (LCD)~\cite{angeli2008fast,angeli2008realtime,fpr2012,tsintotas2018assigning}, and global localization~\cite{wang2006coarsetofine,newman2006outdoor}. It aims to predict the localization where the robotic or agent arrives. A flurry of recent efforts in deep learning have been dedicated to the vision-based approaches for the place recognition task. With the development of convolutional neural networks (CNNs)~\cite{alexnet2012,vgg2014,googlenet2015,resnet2016}, the vision-based methods have achieved promising results with 2D images on the place recognition task. However, vision-based methods often fail in some cases, such as different lighting, seasonal changes, and different climates. Therefore, it is difficult to correctly match the pair of images for place recognition under these circumstances. To tackle these issues, 3D point clouds provide another feasible solution for place recognition, which is invariant to different lighting conditions, seasonal changes, and etc. 

Recently, PointNet~\cite{qi2017pointnet} has once again aroused interest in 3D point cloud analysis. PointNet provides an efficient point cloud feature learning by simply using max pooling function and multi-layer perceptrons (MLPs) to construct the network. Lately, many efforts~\cite{qi2017pointnet++,wang2018dynamic,pointcnn2018,zhao2019pointweb,pointconv2019} are made on point cloud feature learning. Based on PointNet, Mikaela~\emph{et al.} first proposed PointNetVLAD~\cite{pointnetvlad2018} for large-scale place recognition, which is a simple combination of the existing PointNet and NetVLAD~\cite{netvlad2016}. However, it ignores the local features of point clouds, which are effective for point cloud feature learning. Therefore, to enhance the local feature learning, PCAN~\cite{pcan2019} proposed a point contextual attention network that predicts the importance of each local feature, thereby improving the discrimination of point cloud features. Compared to PointNetVLAD, although PCAN can improve performance on place recognition, it requires a lot of computation resources and long inference time. Lately, Liu~\emph{et al.} proposed LPD-Net~\cite{lpdnet2019} for place recognition that takes advantage of handcrafted features and designs a graph-based aggregation module, thereby greatly improving the performance. Nonetheless, LPD-Net also suffers from large resource consumption. Therefore, how to improve performance and reduce resource consumption is still a challenging problem for the place recognition task.

To tackle the aforementioned problems, we propose an efficient point cloud learning network (EPC-Net) for point cloud based place recognition. Specifically, EPC-Net contains two subnetworks: proxy point convolutional neural network (PPCNN) and grouped VLAD network (G-VALD). The proxy point convolutional neural network focuses on extracting multi-scale local geometric features, whereas the G-VLAD network aims to generate discriminative global descriptors from the obtained multi-scale local geometric features. In order to reduce the inference time for efficient retrieval, inspired by edge convolution (EdgeConv)~\cite{wang2018dynamic}, we design a lightweight but effective module, called ProxyConv, to aggregate the local geometric features of point clouds. Unlike EdgeConv, we construct the static $k$-nearest neighbor graph in the spatial space rather than the dynamic $k$-nearest neighbor graph in the feature space. We further use an adjacent matrix to describe the static $k$-nearest neighbor graph in the spatial space. The points are close to each other in the feature space does not mean that they are close in the spatial space. Therefore, using the spatial space to construct the graph can better capture the local geometric structures of point clouds. Furthermore, we use the proxy point to replace the $k$-nearest neighbors used in the edge convolution. The proxy point is generated by averaging neighbors and can be achieved by multiplying the spatial neighbor matrix and the point feature. Therefore, the memory consumption of $k$ neighbors is reduced to that of one proxy point. By stacking the ProxyConv modules, we can construct the proxy point convolutional neural network (PPCNN) and obtain multi-scale local geometric features. Note that in PPCNN, since all ProxyConv modules share the same spatial adjacent matrix, it can reduce resource consumption and inference time. After that, we propose a grouped VLAD network (G-VLAD) to aggregate global descriptors for retrieval. Since the original VLAD network simply uses one fully connected layer to map the very high-dimensional global vector to the fixed dimensional vector for retrieval, it suffers from a large number of parameters and consumes a lot of memory. To alleviate this issue, we propose a grouped fully connected (GFC) layer to decompose the high-dimensional vector into a group of relative low-dimensional vectors and adopt fully connected layers to map them to new feature vectors. Finally, we obtain the global descriptor by summing all feature vectors. We construct G-VLAD network by integrating the GFC layer into the VLAD network. As a result, we can obtain discriminative global descriptors and reduce the number of parameters of the network greatly. To further reduce the inference time, we also develop a simple version of EPC-Net, called EPC-Net-L, which consists of two ProxyConv modules and one max pooling layer to aggregate the global descriptors. By distilling the knowledge from EPC-Net, EPC-Net-L can obtain discriminative global descriptors for retrieval. Extensive experiments have demonstrated that our proposed method can not only achieve the state-of-the-art, but also reduce inference time and resource consumption. In summary, the main contributions of this paper are as follows:

\begin{itemize}
\item We develop an efficient point cloud learning network (EPC-Net) that can achieve the state-of-the-art of point cloud based retrieval.

\item We present a lightweight but effective module, ProxyConv, to characterize the local geometric features of point clouds, which can reduce resource consumption.

\item We design a grouped VLAD network (G-VLAD) to aggregate discriminative global descriptors, where a grouped fully connected (GFC) layer is used to reduce the number of parameters of the network.

\item To further reduce the inference time, we propose a simple version of EPC-Net, called EPC-Net-L, which can obtain discriminative global descriptors by distilling the knowledge from EPC-Net.
\end{itemize}

The rest of the paper is organized as follows: Section~\ref{sec:related_work} introduces related work. In Section~\ref{sec:proposed_approach}, we present our EPC-Net for point cloud based place recognition. Section~\ref{sec:experiments} shows experimental results and Section~\ref{sec:conclusion} concludes the paper.

\section{Related Work}\label{sec:related_work}
\subsection{Handcrafted 3D Descriptors}
Handcrafted features of point clouds are usually used for a variety of 3D applications. The key issue is to form robust descriptors, which can effectively characterize the local geometric structures of point clouds. In the early years, Spin Images~\cite{johnson1999using} computes the histogram features of points distribution in the 2D project. TriSI~\cite{guo2013trisi} is an improvement of Spin Images using the local reference frame (LRF). However, both~\cite{johnson1999using,guo2013trisi} are sensitive to noise. Signature of Histograms of Orientations (SHOT)~\cite{salti2014shot} combines the geometric distribution information and histogram statistical information. Normal Aligned Radial Feature (NARF)~\cite{steder2010narf} is a 3D feature point detection and description algorithm. 
RoPS~\cite{guo2013rotational} is a local feature descriptor for 3D rigid objects based on the rotational projection statistics, but it is sensitive to occlusions and clutter. Point Feature Histograms (PFH)~\cite{rusu2008aligning} representation is based on the relationships between the points in the $k$-neighborhood and their estimated surface normals. In order to reduce computational cost, Fast Point Feature Histograms (FPFH)~\cite{rusu2009fast} is proposed. However, \cite{rusu2008aligning,rusu2009fast} need high data density to support descriptor computation, which makes it impossible to extend to large-scale environments. Most handcrafted descriptors are designed for specific tasks. However, due to the sparseness of the point cloud in the wild environment, these methods are not suitable for point cloud based large-scale place recognition.




\subsection{Learned 3D Descriptors}
With the development of deep learning on 3D vision, more methods are presented to learn the representation of 3D data. Thus, handcrafted features are replaced by deep features, which are more robust in a variety of tasks such as 3D classification, 3D object detection, 3D place recognition, etc. 
Many efforts have been made to obtain discriminative descriptors of different representations. 

At the volumetric representation, methods~\cite{wu20153d,maturana2015voxnet,sedaghat2016,qi2016volumetric} usually voxelize a point cloud into 3D grids, and then apply a 3D convolution neural network (3D CNN) to learn descriptors for object recognition and classification. Wu~\emph{et al.}~\cite{wu20153d} introduced ShapeNets to learn descriptors of 3D shapes, which are represented as the voxel grids. Maturana~\emph{et al.}~\cite{maturana2015voxnet} proposed VoxNet to obtain robust descriptors for 3D object recognition. However, the voxelized representation is not always effective because it represents both the occupied and non-occupied parts of the scene, resulting in a huge unnecessary need for memory storage. To alleviate it, OctNet~\cite{octnet2017} considered using a set of unbalanced octrees to hierarchically partition space by exploiting the sparsity of the input data. However, the voxelization step inherently introduces discretization artifacts and information loss. Moreover, high-resolution voxels will result in higher memory consumption, while low resolution will result in loss of details.

View-based methods~\cite{su15mvcnn,Su20153DAssistedFS,qi2016volumetric,zhao2017multiview,yang2019learning} represented 3D data by projecting it into a 2D view, and then use classic convolution neural network (CNN) to extract multi-view features from collected multi-view 2D images to learn discriminative descriptors for 3D shape recognition. MVCNN~\cite{su15mvcnn} is the first work that aggregates multi-view features as a global descriptor, but visual features are not stable for large view point changes. Therefore, Su~\emph{et al.}~\cite{Su20153DAssistedFS} proposed robust descriptors for object recognition by synthesizing its features for other views of the same object. Furthermore, Qi~\emph{et al.}~\cite{qi2016volumetric} utilized both volumetric CNNs and multi-view CNNs to better capture the global descriptors for 3D object classification. Lately, Yang~\emph{et al.}~\cite{yang2019learning} exploited the relation of region-region and view-view over a group of views, and aggregated these views to obtain a discriminative descriptor for 3D object. However, view-based methods are sensitive to 3D data density. For example, view-based methods will result in loss of detail for sparse 3D point clouds in the wild environment.

Point-based representation is an efficient representation because it directly uses raw 3D data without any post-processing and the point-based methods have lower memory consumption. Qi~\emph{et al.}~\cite{qi2017pointnet} proposed PointNet, which simply uses a symmetric function ($i.e.$, max pooling function) to aggregate a global descriptor of 3D point cloud for 3D classification and segmentation. In order to improve the discrimination of descriptors, Qi~\emph{et al.}~\cite{qi2017pointnet++} applied max pooling function to the local neighborhoods to aggregate local descriptors and proposed PointNet++. Lately, many efforts~\cite{wang2018dynamic,pointcnn2018,pointconv2019,zhao2019pointweb,thomas2019kpconv} have been made to improve the discrimination of the global descriptors of 3D point clouds. Compared with voxel-based and view-based methods, point-based methods can also achieve amazing results, while having lower memory consumption, and can be easily extended to large-scale environments. 

\subsection{Large-Scale Place Recognition}
Many efforts~\cite{irschara2009structure,cieslewski2016point,dube2017segmatch,pointnetvlad2018,pcan2019,sun2019point,lpdnet2019,dh3d2020} have been introduced for point cloud based large-scale place recognition. Specifically, \cite{irschara2009structure} proposed a place recognition method based on structure from motion point clouds and also proposed a compressed 3D scene representation to improve recognition rates. Lately, \cite{cieslewski2016point} proposed a structural descriptor that aggregated sparse visual information into a compact representation to provide a discriminative descriptor for place recognition. Based on 3D segmentation, \cite{dube2017segmatch} proposed SegMatch, a place recognition algorithm that combined the advantages of local and global descriptions while reducing their individual drawbacks. \cite{sun2019point} proposed a point cloud based place recognition system that adopted a convolution neural network pre-trained on color images to extract deep features rather than using handcrafted features. However, the aforementioned methods are complicated and not easy to employ. Recently, \cite{pointnetvlad2018} proposed PointNetVLAD, a lightweight network, to deal with the point cloud based large-scale place recognition and achieved good results. Specifically, based on PointNet~\cite{qi2017pointnet}, it explicitly leverages the existing NetVLAD~\cite{netvlad2016} to improve the discrimination of the global descriptor for place recognition. After that, PCAN~\cite{pcan2019} introduced an attention module based on \cite{pointnetvlad2018} to encode the local features into a discriminative global descriptor, where a point contextual network is used to aggregate multi-scale context information to obtain an attention map. 
However, PointNetVLAD and PCAN only consider the single point features and greatly ignore the local features of point clouds. 
To tackle it, LPD-Net~\cite{lpdnet2019} utilized a graph-based neighborhood aggregation module to aggregate local features and then adopted NetVLAD to obtain global descriptors for place recognition. 
Note that the input of LPD-Net is the raw point clouds plus extra handcrafted features. Computing handcrafted features are very time-consuming, so it is difficult to apply it to real-world testing. In addition, it suffers from a large number of parameters and higher memory consumption due to local feature learning. Likewise, DH3D~\cite{dh3d2020} proposed a siamese network that jointly learns 3D local feature detection and description. For retrieval, DH3D adopted the attention-based VLAD network proposed in~\cite{pcan2019} to obtain discriminative global descriptors. Thus, due to the attention-based VLAD network, DH3D still faces a huge computation cost. 
In this paper, we present an efficient point cloud learning network that can efficiently extract local features of point cloud and produce a discriminative global descriptor for place recognition. What's more, our method enjoys the characteristics of low resource consumption and fast inference time.









\begin{figure*}
	\centering
	\includegraphics[width=0.98\textwidth]{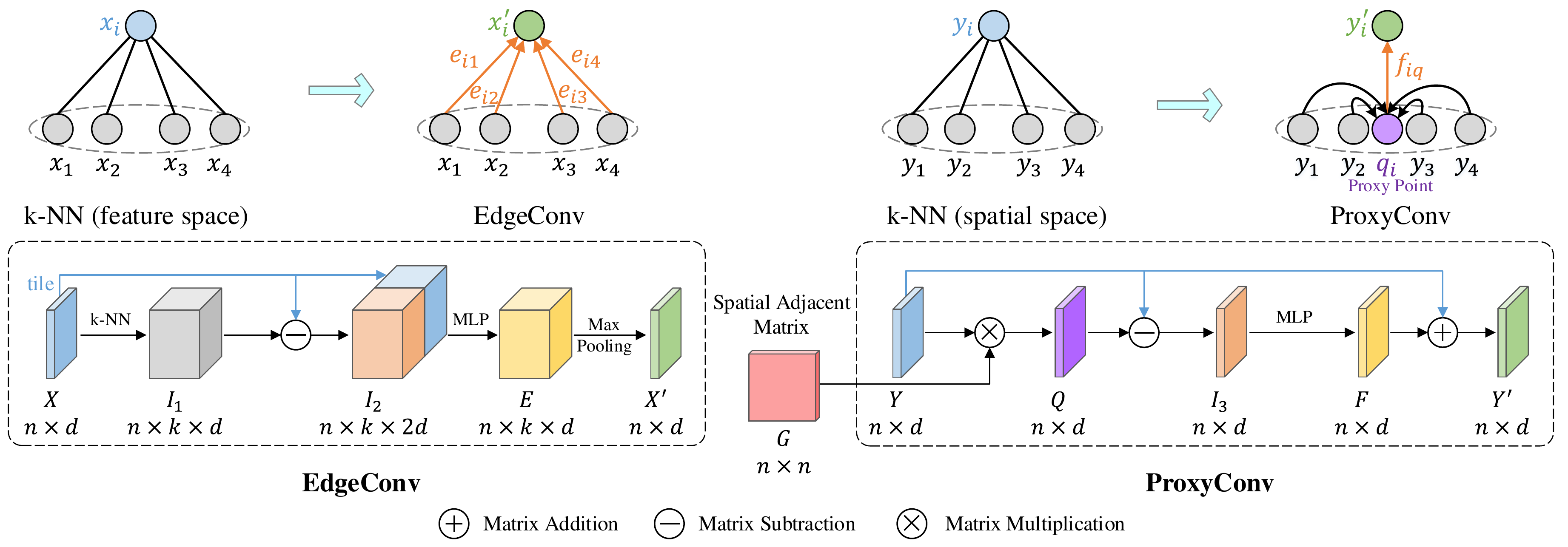}
	\caption{{\bf Left}: The architecture of EdgeConv~\cite{wang2018dynamic} module. {\bf Right}: The architecture of our ProxyConv module. The ProxyConv module is a lightweight version of the EdgeConv module.
	}
	\vspace{-5pt}
	\label{fig:fastedgeconv}
\end{figure*}

\section{PROPOSED METHOD}\label{sec:proposed_approach}
In this section, we introduce the details of our method for point cloud based retrieval. In Section~\ref{sec:sgcnn}, we first describe the proxy point convolutional neural network (PPCNN) for extracting local geometric features. Then, we design a grouped VLAD network (G-VLAD) to aggregate global descriptors in Section~\ref{sec:G-VLAD}. After that, we introduce the architecture of the efficient point cloud learning network (EPC-Net) for point cloud based retrieval in Section~\ref{sec:framework}. To further reduce the inference time, in Section~\ref{sec:distillation}, we propose a simple version of EPC-Net, called EPC-Net-L, which distills the knowledge from EPC-Net to improve the generalization ability.

\subsection{Proxy Point Convolutional Neural Network}\label{sec:sgcnn}
We propose a proxy point convolutional neural network (PPCNN) to aggregate the multi-scale local geometric features. It can not only aggregate the local geometric features of point clouds, but also reduce the inference time, parameters, and resource consumption. PPCNN is constructed by stacking a set of ProxyConv modules, which is the key to aggregating the local geometric features. Our ProxyConv module is a lightweight but effective version of EdgeConv~\cite{wang2018dynamic}. As shown in Fig.~\ref{fig:fastedgeconv}, we present the detailed structure of EdgeConv and ProxyConv modules.

EdgeConv is proposed to aggregate the local geometric features. Let us consider a $d$-dimensional point cloud with $n$ points, denoted by $\bm{X}\in\mathbb{R}^{n\times d}$, where $n$ denotes the number of points and $d$ represents the number of channels of the point features. It first constructs the $k$-nearest neighbor graph of $\bm{X}$ in the feature space. Then, the edge feature is defined by:
\begin{equation}
\bm{e}_{ij} = \text{ReLU}(h_{\Theta}(\bm{x}_i,\bm{x}_j - \bm{x}_i))
\label{equ:edge_convolution}
\end{equation}
where $\bm{x}_i$ is regarded as the central point and $\bm{x}_j$ is one of the $k$-nearest neighbors. Here, $h_{\Theta}:\mathbb{R}^d\times\mathbb{R}^d\rightarrow\mathbb{R}^d$ is a nonlinear function with learnable parameters $\Theta$. The local neighborhood information is captured by $\bm{x}_j-\bm{x}_i$. Finally, a channel-wise symmetric function, max pooling, is applied to aggregate the local geometric features: $\bm{x}^{\prime}_i=\max_{j}e_{ij}$. From the figure, the memory consumption of EdgeConv is $O(n*d)$+$O(n*k*d)$+$O(n*k*2d)$+$O(n*k*d)$+$O(n*d)$=$O(n*(2+4k)*d)$. If we stack $m$ EdgeConv modules, the total memory consumption is up to $O(m*n*(2+4k)*d)$. When increasing the number of channels $d$ and neighbors $k$, the memory consumption will increase rapidly. In addition, EdgeConv needs to recalculate the $k$-nearest neighbor graph in the feature space, which will increase the computational cost, thereby increasing the inference time. LPD-Net~\cite{lpdnet2019} adopts EdgeConv to extract features. Therefore, it takes up a lot of memory and requires a long inference time for retrieval.

To alleviate this issue, we consider two points in our ProxyConv module: (1) we construct the $k$-nearest neighbor graph in the spatial space; (2) we use the proxy point to replace the $k$-nearest neighbors used in the edge convolution. For the first point, since the $k$-nearest neighbors are close to each other in the feature space does not mean that they are close to each other in the spatial space, we construct the $k$-nearest neighbor graph in the spatial space to better characterize the local geometric structure. Moreover, the $k$-nearest neighbor graph in the spatial space is a static graph, which only needs to be calculated once, thereby saving the computational cost. For the second point, when calculating the local geometric information, we replace the $k$-nearest neighbors with the proxy point. In this way, the memory consumption of the feature map will be reduced from $O(n*k*d)$ (see $\bm{I}_1$ in Fig.~\ref{fig:fastedgeconv}) to $O(n*d)$ (see $\bm{A}$ in Fig.~\ref{fig:fastedgeconv}). In this paper, the proxy point is generated by averaging the features of the $k$-nearest neighbors. In our experiments, we found that the proxy point obtained by averaging the features of local neighbors can work well. Moreover, this average operation can be achieved by multiplying the point feature with the spatial adjacent matrix. Therefore, we can share the spatial adjacent matrix in all ProxyConv modules to reduce memory consumption and computational cost.

The detailed structure of the ProxyConv module is presented in the right part of Fig.~\ref{fig:fastedgeconv}. First of all, we introduce the spatial adjacent matrix $\bm{G}\in\mathbb{R}^{n\times n}$. An adjacent matrix is a square matrix used to represent a graph. The elements of the matrix indicate whether the pairs of points are adjacent or not in the graph. In this paper, we construct the adjacent matrix $\bm{G}$ according to the $k$-nearest neighbor ($k$-NN) graph in the spatial space. We first compute the Euclidean distance between points $i$ and $j$ by:
\begin{equation}
D_{ij} = \|\bm{p}_i - \bm{p}_j\|^{2}
\label{equ:euclidean_distance}
\end{equation}
where $\bm{p}_i\in\mathbb{R}^{3}$ and $\bm{p}_j\in\mathbb{R}^{3}$ are two vectors of coordinates. The $k$-nearest neighbors of each point can be regarded as the $k$-smallest elements in each row in $\bm{D}$. To obtain the spatial adjacent matrix $\bm{G}\in\mathbb{R}^{n\times n}$, we binarize the elements in each row: the $k$-smallest elements are set to one, and the rest elements are set to zero. After obtaining the spatial adjacent matrix, we can use it to generate proxy points. Formally, the formula for averaging local neighbor features through matrix multiplication is written as:
\begin{equation}
\bm{Q} = \frac{1}{k}\bm{G}\times\bm{Y}
\label{equ:average_pooling_matrix}
\end{equation}
where $\bm{Y}\in\mathbb{R}^{n\times d}$ is the point feature and $\bm{Q}=\{\bm{q}_1,\cdots,\bm{q}_n\mid \bm{q}_i\in\mathbb{R}^{d}\}$ is the generated new proxy points, where $\bm{q}_i$ is the point feature of $i$-th proxy point. Therefore, the original feature map $\bm{I}_1\in\mathbb{R}^{n\times k\times d}$ of EdgeConv can be reduced to $\bm{Q}\in\mathbb{R}^{n\times d}$ of ProxyConv in Fig.~\ref{fig:fastedgeconv}.

\begin{figure*}
	\centering
	\includegraphics[width=1.0\textwidth]{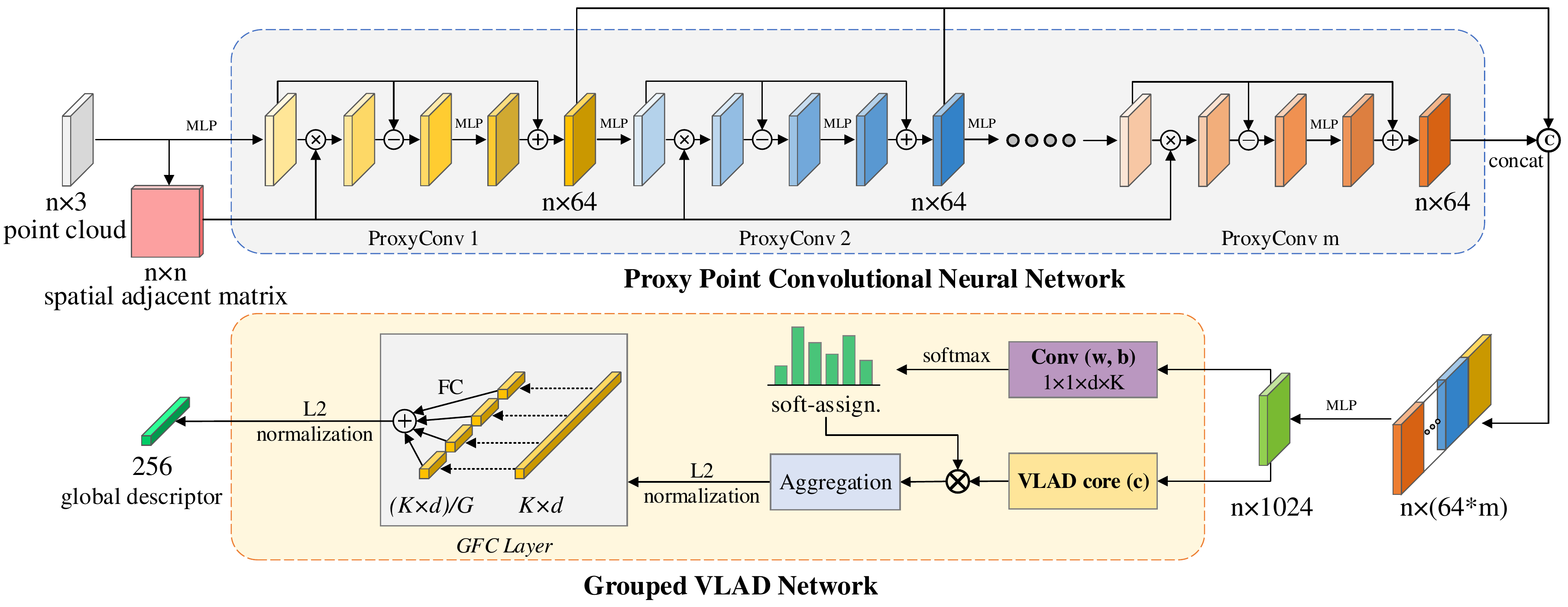}
	\vspace{-20pt}
	\caption{The architecture of our efficient point cloud learning network (EPC-Net) for point cloud based place recognition. Given the raw point clouds, we first compute the spatial adjacent matrix in the spatial space. Then, we use the proxy point convolutional neural network (PPCNN) to extract multi-scale local geometric features. After that, we concatenate the multi-scale feature maps of each module to obtain a (64*$m$)-dimensional feature map. Next, we use multi-layer perception (MLP) to map it into a 1024-dimensional feature space. Finally, we use the proposed grouped VLAD network (G-VLAD) to obtain a 256-dimensional global descriptor, in which the grouped fully connected (GFC) layer is used to reduce the number of parameters. Note that the spatial adjacent matrix only needs to be calculated once in the entire network.}
	\vspace{-10pt}
	\label{fig:framework}
\end{figure*}

As a result, we can use the central point and proxy point to calculate the local geometric information instead of using the central point and its neighbors. Therefore, the new aggregated feature is defined by:
\begin{equation}
\bm{f}_{iq} = \text{ReLU}(g_{\Theta}(\bm{q}_i-\bm{y}_i))
\label{equ:fast_edge_convolution}
\end{equation}
where $\bm{f}_{iq}\in\mathbb{R}^{d}$ is the generated new feature and $g_{\Theta}:\mathbb{R}^d\rightarrow\mathbb{R}^d$ is a nonlinear function with learnable parameters $\Theta$. Note that in Eq.~(\ref{equ:edge_convolution}), if $i$ is equal to $j$, then $\bm{x}_j-\bm{x}_i$ is zero. Therefore, the central point feature $\bm{x}_i$ should be concatenated in Eq.~(\ref{equ:edge_convolution}). In Eq.~(\ref{equ:fast_edge_convolution}), $\bm{q}_i$ is not equal to $\bm{y}_i$, so we do not concatenate $\bm{y}_i$ to $\bm{q}_i-\bm{y}_i$. Finally, we define the ProxyConv by adding the input point feature on the generated new feature: $\bm{y}_i^{\prime}=\bm{f}_i+\bm{y}_i$. From Fig.~\ref{fig:fastedgeconv}, the memory consumption of ProxyConv is $O(n*5d)$+$O(n*n)$. If we stack $m$ ProxyConv modules, the total memory consumption is up to $O(m*n*5d)+O(n*n)$. Since the adjacent matrix is shared in all ProxyConv modules, the memory consumption of the adjacent matrix is calculated only once. The memory consumption ratio of the ProxyConv and EdgeConv modules is written as:
\begin{equation}
\begin{aligned}
R(\text{PC}, \text{EC}) &= \frac{m*n*5d+n*n}{m*n*(2+4k)*d}\\
&=\frac{5}{2+4k}+\frac{n}{m*(2+4k)*d}
\end{aligned}
\label{equ:ratio}
\end{equation}
where $\text{PC}$ and $\text{EC}$ represent the ProxyConv and the EdgeConv, respectively. Generally, the values of $n$, $k$, $d$ and $m$ are 4096, 20, 64, and 4, respectively. The ratio of Eq.~(\ref{equ:ratio}) is about 25\%. Moreover, thanks to the spatial adjacent matrix, the memory consumption of the ProxyConv is independent of the number of neighbors. However, as the number of neighbors increases, the memory consumption of EdgeConv also increases. Moreover, it can be seen from Eq.~(\ref{equ:edge_convolution}) and Eq.~(\ref{equ:fast_edge_convolution}), our ProxyConv has smaller number of parameters. 

To construct the proxy point convolutional neural network (PPCNN), we directly stack the ProxyConv module. As shown in the top part of Fig.~\ref{fig:framework}, we provide the detailed structure of our PPCNN. Since we construct the spatial adjacent matrix in the spatial space, all modules can share the same spatial adjacent matrix. Therefore, the spatial adjacent matrix only needs to be calculated once in the entire PPCNN. By stacking the ProxyConv modules, we can obtain multi-scale local features. Thanks to our ProxyConv module, our proxy point convolutional neural network can have low memory consumption and short inference time.

As aforementioned, we use one proxy point to replace the $k$-nearest neighbors. Although it seems to lose some local information, we want to emphasize memory consumption and inference time in point cloud based place recognition. Surprisingly, compared with the original EdgeConv, by using proxy points, we can approximately reduce 75\% of memory consumption. Likewise, due to the reduction of memory computation and feature map size, the inference time of the network is also reduced. In addition, the proxy point is generated by averaging the point features of the local neighborhood. In other words, the proxy point is a compound point that aggregates local information. Therefore, using proxy points can not only reduce memory, but also utilize the local information of point clouds.

\subsection{Grouped VLAD Network}\label{sec:G-VLAD}
Once we have obtained the local descriptors from the efficient proxy point convolutional neural network, we extract global descriptors for point cloud based place recognition. Based on NetVLAD~\cite{netvlad2016}, we design a grouped VLAD network (G-VLAD), which can obtain discriminative global descriptors with fewer parameters.

Given a set of learned local feature descriptor $\{\bm{f}_i\mid i=1,2,\cdots,n\}$, where $n$ is the number of points and $\bm{f}_i\in\mathbb{R}^{d}$ is $d$-dimensional descriptor vector. The original VLAD network first learns $K$ cluster centers $\{\bm{c}_1,\cdots,\bm{c}_K\mid\bm{c}_k\in\mathbb{R}^{d}\}$, and then compute the subvector $\bm{V}_k$ for each cluster center $\bm{c}_k$. Specifically, $\bm{V}_k$ can be viewed as the weighted sum of the differences between $\bm{f}_i$ and $\bm{c}_k$, and is written as:
\begin{equation}
\bm{V}_{k}=\sum\nolimits_{i=1}^{n}a^{k}_{i}\left(\bm{f}_{i}-\bm{c}_{k}\right)
\label{equ:vlad}
\end{equation}
where $\bm{V}_{k}\in\mathbb{R}^{d}$, and $a^{k}_{i}$ is a soft weight. The output global descriptor vector $\bm{V}=[\bm{V}_1,\cdots,\bm{V}_K]$ is the aggregation of the local feature vectors, where $\bm{V}\in\mathbb{R}^{K\times d}$. When $K$ or $d$ becomes larger, the obtained global descriptor is a very high-dimensional vector, $i.e.$, ($K\times d$)-dimensional vector, which is computationally expensive in terms of the time and resource during retrieval. Generally, the ($K\times d$)-dimensional vector is at least a $10^5$-dimensional vector in point cloud based retrieval. Therefore, the fully connected (FC) layer is usually used to compress the descriptor vector from the very high-dimensional vector to a compact low-dimensional vector. However, due to the high-dimensional input vector, the last FC layer will cause a large number of parameters and memory consumption.

To address this problem, we introduce the grouped fully connected (GFC) layer to decompose a very high-dimensional vector into a group of relatively low-dimensional vectors so that the number of parameters of the fully connected layer can be reduced and the discrimination ability of the fully connected layer can be maintained. Specifically, as shown in Fig.~\ref{fig:framework} (refer to grouped VLAD network), we first divide the ($K\times d$)-dimensional vectors into $G$ groups, $i.e.$, $G=[\bm{U}_1,\cdots,\bm{U}_G\mid\bm{U}_g\in\mathbb{R}^{(K\times d)/G}]$. Note that the dimension of each group is $(K\times d)/G$, where $G$ is constrained to be divisible by $K\times d$. Consequently, the fully connected layers are separately performed within each group to obtain new vectors $\bm{U}_g^{'}\in\mathbb{R}^{\mathcal{O}}$. In order to aggregate the final global descriptor, here we sum all the grouped new vectors. The aggregated vector is written as:
\begin{equation}
\bm{\mathcal{X}}=\sum\nolimits_{g=1}^{G}\mathcal{FC}(\bm{U}_g)=\sum\nolimits_{g=1}^{G}\bm{U}_g^{'}
\label{equ:gfc}
\end{equation}
where $\mathcal{FC}: \mathbb{R}^{(K\times d)/G}\rightarrow \mathbb{R}^{\mathcal{O}}$, and $\mathcal{X}\in\mathbb{R}^{\mathcal{O}}$ represents the global descriptor for place recognition. It can be found that the number of parameters $K\times d\times\mathcal{O}$ is reduced to $(K\times d\times\mathcal{O})/G$. In the experiment, we empirically set $G=$ 8, which means that we can reduce the parameters by 8 times. Note that the GFC layer is a generalization of the FC layer as with $G$ = 1 the two are equivalent. We observe that the generalization ability of the VLAD network is hardly affected by the GFC layer. Due to the redundancy of the high-dimensional vector obtained from the original VLAD, the decomposed low-dimensional vector can discriminatively characterize the manifold structure of the high-dimensional feature space of the VLAD descriptor. 

\subsection{Network Architecture}\label{sec:framework}
As illustrated in Fig.~\ref{fig:framework}, we present the architecture of our efficient point cloud learning network (EPC-Net) for point cloud based place recognition. The framework mainly consists of two parts: proxy point convolutional neural network (PPCNN) and grouped VLAD network (G-VLAD). Given a raw point clouds with only spatial coordinates ($i.e.$, XYZ coordinates), we use the proposed proxy point convolutional neural network to extract multi-scale local geometric features. Then, we concatenate the multi-scale feature maps to obtain a (64*$m$)-dimensional feature map, where $m$ is the number of stacked modules. After that, we apply MLP to map the obtained feature into 1024-dimensional feature map. Note that in PPCNN, since the adjacent matrix only needs to be calculated once in all $m$ ProxyConv modules, so it saves a lot of computational time and memory consumption. Since we have obtained the local feature of each point, we leverage the proposed grouped VLAD network to aggregate the global descriptor for retrieval. G-VLAD network takes the feature map of the entire point clouds as input and outputs a vector as a global descriptor. In the G-VLAD network, the grouped fully connected (GFC) layer is used to reduce the number of parameters during training. For the place recognition task, we compute the distance in the descriptor space between the test scene and the query scene to determine whether they are the same places. Compared with the previous methods~\cite{pointnetvlad2018,lpdnet2019,pcan2019,dh3d2020}, our EPC-Net can not only efficiently aggregate local geometric features and obtain discriminative global descriptors, but also reduce the memory consumption, parameters, and inference time.

In order to reduce the inference time, we also construct a simple version of EPC-Net, called EPC-Net-L. We directly stack two ProxyConv modules to construct the lightweight proxy point convolutional neural network. Thus, we can obtain a 128-dimensional feature map by concatenating two 64-dimensional feature maps. To further reduce the inference time, we only use max pooling to aggregate the global descriptors for final retrieval, instead of using the proposed G-VLAD network. Therefore, the number of parameters of the EPC-Net-L is smaller than that of EPC-Net. Although we can obtain a shorter inference time, the performance of EPC-Net-L is much lower than EPC-Net. In the next subsection, we employ distilling knowledge from EPC-Net to improve the performance of EPC-Net-L.

\subsection{Network Distillation}\label{sec:distillation}
Our end-goal is to develop an efficient network that can reduce the inference time and still achieve good performance. To this end, we propose a teacher-student network, in which the teacher network is the proposed efficient point cloud learning network (EPC-Net) mentioned above while the student network is a simple version of EPC-Net, called ECP-Net-L.

\subsubsection{Teacher Network} As shown in Fig.~\ref{fig:framework}, we use the proposed EPC-Net as the teacher network, which integrates the proxy point convolutional neural network and the grouped VLAD network for extracting the discriminative global descriptors for point cloud based place recognition. Since we focus on designing an efficient network for place recognition, we adopt the same lazy quadruplet loss used in PointNetVLAD~\cite{pointnetvlad2018} to train the network. Given a point cloud, we construct the quadruples $\mathcal{T}=\left(P_a, P_{pos}, \{P_{neg}\}, P_{{neg}^{*}}\right)$, in which $P_a$, $P_{pos}$, $\{P_{neg}\}$ and $P_{{neg}^{*}}$ represent the anchor sample, the positive sample similar to the anchor, a set of negative samples dissimilar to the anchor, and the randomly sampled sample from the training dataset, respectively. The lazy quadruplet loss is written as:
\begin{equation}
\begin{aligned}
\mathcal{L}_{lazyQuad} &=\max _{j}\left(\left[\alpha+\delta_{pos}-\delta_{neg}\right]_{+}\right) \\
&+\max_{k}\left(\left[\beta+\delta_{ {pos}}-\delta_{{neg}_{k}^{*}}\right]_{+}\right)
\label{equ:lazyquad}
\end{aligned}
\end{equation}
where $[\cdots]_{+}$ indicates the hinge loss and $\alpha/\beta$ are the given margins. Same as PointNetVLAD, we set $\alpha=$ 0.5 and $\beta=$ 0.2 in teacher network. In addition, $\delta_{pos}$, $\delta_{neg}$ and $\delta_{{neg}_{k}^{*}}$ are the distances between the corresponding sample pairs, respectively.

\subsubsection{Student Network} In addition to the teacher network, the student network is a simple version of EPC-Net. Specifically, we only use two ProxyConv modules to extract local geometric features. After that, a global max pooling is used to get the point cloud global feature, followed by the fully connected layers, which are used to transform the high-dimensional vector to a low-dimensional vector, $i.e.$, the global descriptor. Compared with the previous methods~\cite{pointnetvlad2018,pcan2019,lpdnet2019,dh3d2020}, the student network has a simpler structure, and therefore it has few parameters and short inference time. Since both the teacher network and student network have similar feature extraction modules, we leverage the teacher network to guide the training of the student network to improve its generalization ability. Suppose that the learned global descriptors are $\bm{\mathcal{X}}_S$ and $\bm{\mathcal{X}}_T$ for the student network and the teacher network, respectively. We encourage the global descriptor learned by the teacher network and the global descriptor learned by the student network to be similar. Specifically, we use the sum squared error (SSE) loss to achieve this goal, and the formula is written as:
\begin{equation}
\mathcal{L}_{sse}=\left\|\bm{\mathcal{X}}_T-\bm{\mathcal{X}}_S\right\|^{2}
\end{equation}
where $\bm{\mathcal{X}}_S\in\mathbb{R}^{\mathcal{O}}$ and $\bm{\mathcal{X}}_T\in\mathbb{R}^{\mathcal{O}}$. Note that in this way, the student network can be guided to learn useful information from the teacher network for efficient retrieval on point clouds. In addition, we also use the same lazy quadruplet loss as in Eq.~(\ref{equ:lazyquad}) to train the student network. Consequently, the final loss for training the student network is defined as:
\begin{equation}
\mathcal{L}_{final}=\mathcal{L}_{lazyQuad}+\lambda\mathcal{L}_{sse}
\end{equation}
where $\lambda$ are the hyperparameters and we set $\lambda=$ 0.1 in the experiment.

\begin{figure*}[t]
	\centering
	\includegraphics[width=1.0\textwidth]{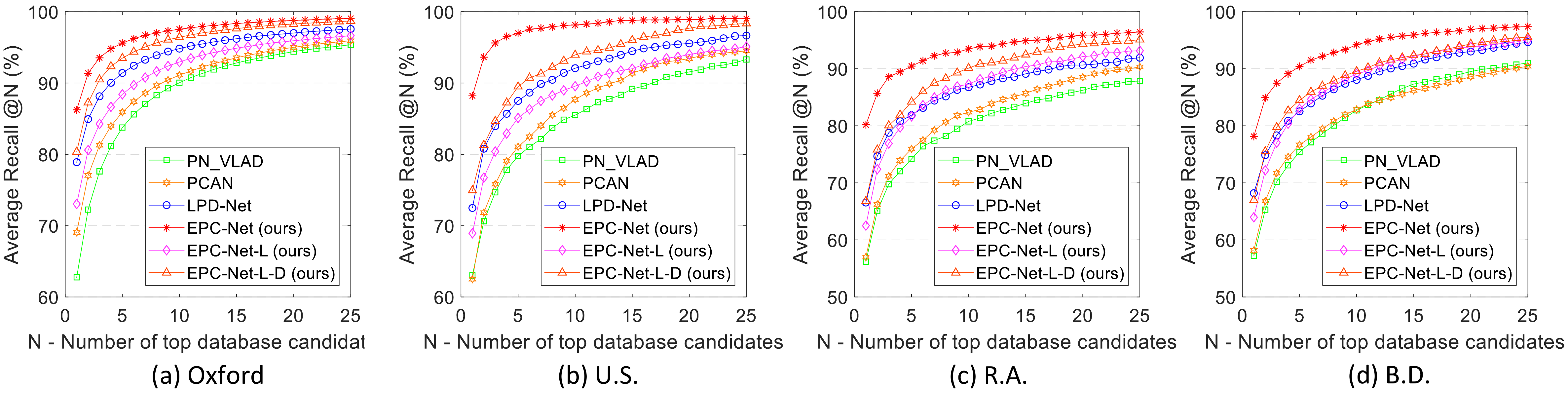}
	\vspace{-20pt}
	\caption{Average recall of different methods on the Oxford, U.S., R.A., and B.D. datasets, respectively. Note that all methods are trained on the train set of the Oxford dataset and tested on the other four datasets.}
	\vspace{-10pt}
	\label{fig:comparison_curve}
\end{figure*}

We first train the teacher network with the lazy quadruplet loss $\mathcal{L}_{lazyQuad}$. Then, we train the student network with the integrated loss $\mathcal{L}_{final}$. Note that during the training of the student network, we froze the parameters of the trained teacher network, that is, it only performs forward inference.

\section{Experiments}\label{sec:experiments}
\subsection{Experimental Settings}
\subsubsection{Datasets}
We adopt the benchmark datasets proposed in PointNetVLAD~\cite{pointnetvlad2018} for LiDAR-based place recognition in the experiment. The datasets are built based on the open-source Oxford RobotCar~\cite{maddern20171} and three in-house datasets of a university sector (U.S.), a residential area (R.A.) and a business district (B.D.). Specifically, LiDAR data of four regions is collected from the LiDAR sensor mounted on a car. Note that in practice, PointNetVLAD removes the non-informative ground planes and downsamples the LiDAR data to 4096 points. In addition, each downsampled submap is tagged with a UTM coordinate. To produce the annotation of each submap, PointNetVLAD adopts the point clouds at most 10m apart as the positive pairs and the point clouds at least 50m apart as the negative pairs. For all submaps collected from the Oxford dataset, the 21,711 submaps are used for training and the rest 3030 submaps are used for testing. The submaps of the in-house dataset are all used to test the generalization of the methods trained only on the Oxford dataset. Note that in the evaluation scheme, if the distance between the query submap and the retrieved submap is smaller than 25m, it is regarded that the query point clouds have been successfully located.

\subsubsection{Evaluation metric}
We adopt the most common metric, $i.e.$, recall rate, for evaluating the LiDAR-based place recognition. It usually finds top-$K$ candidates, where $K=1,2,3,\cdots$. Following~\cite{pointnetvlad2018,pcan2019,lpdnet2019}, we use recalls at top 1\%  (@1\%) and at top 1 (@1) for ease of comparison with these methods. Moreover, we also compare our method with the previous methods in the number of parameters, floating point operations (FLOPs), and runtime per frame.

\subsubsection{Implementation details} Our efficient point cloud learning network (EPC-Net) mainly consists of a proxy point convolutional neural network (PPCNN) and a grouped VLAD network (G-VLAD). The architecture of our framework is shown in Fig.~\ref{fig:framework}. In this paper, we use the same LiDAR data of 4096 points as~\cite{pointnetvlad2018}. Specifically, we first stack $m=$ 4 ProxyConv modules, each of which has a neuron size of 4096$\times$64. In addition, the number $k$ of local neighbors is set to 20 for calculating the spatial adjacent matrix. Then, we concatenate the multi-scale feature maps of four modules to obtain a 4096$\times$256 feature map. After that, we use multi-layer perceptron (MLP) to map the 4096$\times$256 feature map into a high-dimensional feature map with the size of 4096$\times$1024. Then, the obtained 4096$\times$1024 feature map is sent to the G-VLAD network to generate a global descriptor for retrieval. In the G-VLAD network, the number $K$ of clusters is 64, and the output dimension $\mathcal{O}$ is 256. In addition, the number $G$ of groups in the GFC layer is 4. For network distillation, we only stack two ProxyConv modules as a lightweight feature extraction network, and use the max pooling to aggregate global descriptors for retrieval. Similarly, we concatenate the multi-scale feature maps of two modules to obtain a 4096$\times$128 feature map, and use MLP to map it to a high-dimensional feature map with the size of 4096$\times$1024. By applying a max pooling on the 4096$\times$1024 feature map, we obtain the 1024-dimensional global descriptor. Here we use one fully connected layer to compress the 1024-dimensional vector into a 256-dimensional vector for retrieval. In the entire network, we use Leaky ReLU~\cite{xu2015empirical} and batch normalization~\cite{ioffe2015batch} after each layer. In addition, we use Adam~\cite{kingma2014adam} with learning rate 5$\times$10$^{-5}$ to train the network. For the hyperparameters, we set $\alpha$ and $\beta$ of the lazy quadruplet loss to 0.5 and 0.2, respectively. For the final loss, we set the hyperparameters $\lambda=$ 0.1 in the experiment. We use the deep learning platform TensorFlow~\cite{abadi2016tensorflow} to complete our method.

\subsection{Results}
To evaluate our method, we conduct experiments on the benchmark dataset proposed in~\cite{pointnetvlad2018}. We compare our method (EPC-Net) with series of previous methods. The method called PN\_MAX~\cite{pointnetvlad2018} uses a max-pooling to obtain a global descriptor after the original PointNet architecture. PN\_STD~\cite{pointnetvlad2018} uses the trained PointNet in ModelNet~\cite{wu20153d} dataset to see whether the model trained on the ModelNet dataset can be extended to large-scale cases. PN\_VLAD~\cite{pointnetvlad2018} uses the VLAD network to replace the max pooling in PointNet. PCAN~\cite{pcan2019} introduces the point contextual attention network to learn task-relevant features when aggregating the global descriptors. LPD-Net~\cite{lpdnet2019} utilizes both the spatial and feature space through the graph neural network to enhance local features to obtain the discriminative global descriptors. DH3D~\cite{dh3d2020} uses the same attention-based VLAD network as in PCAN, but it uses a different local feature learning network. Note that for all comparison methods, we only use the spatial coordinates of the downsampled point cloud with 4096 points.

\begin{table}[t]
	\centering
	\caption{Comparison results of the average recall (\%) at top 1\% (@1\%) and at top 1 (@1) on the Oxford dataset.}
	\begin{tabular}{l|c|c}
		\hline 
		\hline
		\multicolumn{1}{c|}{Methods} & Ave Recall @1\% & Ave Recall @1 \\ 
		\hline 
		PN\_STD~\cite{pointnetvlad2018}& 46.52 & 31.87 \\ 
		PN\_MAX~\cite{pointnetvlad2018}& 73.87 & 54.16 \\ 
		PN\_VLAD~\cite{pointnetvlad2018} & 81.01 & 62.76 \\ 
		PN\_VLAD refine~\cite{pointnetvlad2018}& 80.71 & 63.33 \\
		PCAN~\cite{pcan2019} & 83.80 & 69.05\\
		PCAN refine~\cite{pcan2019} & 86.04 & 70.72 \\
		LPD-Net$^{*}$~\cite{lpdnet2019} & 90.88 & 80.71 \\
		DH3D~\cite{dh3d2020} & 84.26 & 73.28 \\
		\hline
		EPC-Net (ours) & {\bf 94.74} & {\bf 86.23} \\ 
		EPC-Net-L (ours) & 86.53 & 73.03 \\
		EPC-Net-L-D (ours) & 92.23 & 80.34 \\
		\hline
		\hline
	\end{tabular}
	\vspace{-10pt}
	\label{tab:results_oxford}
\end{table}

\begin{table}
	\centering
	\caption{Comparison results of the parameters and computation required under different networks.}
	\begin{tabular}{l|c|c|c}
		\hline 
		\hline
		\multicolumn{1}{c|}{Methods} & Parameters & FLOPs & Runtime per frame  \\ 
		\hline 
		PN\_VLAD~\cite{pointnetvlad2018} & 19.78M
		& 4.21G
		& 32.94ms \\ 
		PCAN~\cite{pcan2019} & 20.42M
		& 7.73G & 96.94ms\\
		LPD-Net$^{*}$~\cite{lpdnet2019} & 19.81M & 7.80G & 42.80ms\\
		\hline
		EPC-Net (ours) & 4.70M & 3.25G & 32.82ms\\ 
		EPC-Net-L (ours) & {\bf 0.41M} & {\bf 1.37G} & {\bf 25.71ms}\\
		EPC-Net-L-D (ours) & {\bf 0.41M} & {\bf 1.37G} & {\bf 25.71ms}\\
		\hline
		\hline
	\end{tabular}
	\vspace{-10pt}
	\label{tab:results_speed}
\end{table}

\subsubsection{Quantitative results} The quantitative results are shown in Tab.~\ref{tab:results_oxford}, where EPC-Net, EPC-Net-L and EPC-Net-L-D are our methods. Specifically, EPC-Net indicates our efficient point cloud learning network for retrieval. EPC-Net-L is a lightweight version of EPC-Net. To improve the performance, we further propose EPC-Net-L-D, where we use EPC-Net (teacher network) to guide the training of EPC-Net-L (student network). From the table, it can be seen that our EPC-Net significantly outperforms other advanced methods on the Oxford dataset. Our EPC-Net has achieved the state-of-the-art with the average recall of 94.74\% at top 1\%, which proves that our method can generate discriminative global descriptors for retrieval. In addition, our EPC-Net-L is still superior to PN\_VLAD, PCAN, and DH3D with a large margin, which further indicates that our method can also yield good performance in the lightweight version. Since we use EPC-Net to guide the training of EPC-Net-L, the performance of EPC-Net-L-D is better than most of the previous methods. Note that in the table, for a fair comparison, the results of LPD-Net$^*$ is obtained by using their open-source programs and removing the handcrafted features. In addition, the calculation of handcrafted features is very time-consuming, it is difficult to be applied to retrieval in a real environment. With the handcrafted features, the average recall (@1\%) of LPD-Net is reported as 94.92\% in~\cite{lpdnet2019}. Although our EPC-Net is performed without using the handcrafted features, our performance is still comparable with LPD-Net equipped with handcrafted features. Moreover, ``PN\_VLAD refine'' and ``PCAN refine'' represent the results are obtained in the case of using the Oxford dataset plus extra U.S. and R.A. datasets during training. In addition to these two results, other results are obtained just using the Oxford dataset during training. Compared with the methods using the extra datasets, EPC-Net can also achieve impressive performance and improve the average recall (@1\%) from 86.04\% to 94.74\%, a significant gain of 8.7\%. As shown in Tab.~\ref{tab:results_three}, we also test our method in there in-house datasets. Here we report the results obtained by directly testing the model trained only on the Oxford dataset. It can be found that our method also significantly boosts performance than other methods. Furthermore, in Fig.~\ref{fig:comparison_curve}, we also show the recall curves of each method for top 25 retrieval results from the four datasets, where our EPC-Net is superior to other methods. This further demonstrates the effectiveness of our method.

\begin{table}[t]
	\centering
	\caption{Evaluation results of average recall @1\% on the three in-house datasets.}
	\begin{tabular}{l|c|c|c}
		\hline 
		\hline
		\multicolumn{1}{c|}{Methods} & \;\;\;U.S.\;\;\; & \;\;\;R.A.\;\;\; & \;\;\;B.D.\;\;\;  \\ 
		\hline 
		PN-VLAD~\cite{pointnetvlad2018} & 72.63 & 60.27 & 65.30 \\ 
		PCAN~\cite{pcan2019} & 79.05 & 71.17 & 66.81\\
		LPD-Net$^{*}$~\cite{lpdnet2019} & 85.68 & 78.77 & 74.84 \\
		\hline
		EPC-Net (ours) & {\bf 96.52} & {\bf 88.58} & {\bf 84.92} \\ 
		EPC-Net-L (ours) & 82.91 & 76.89 & 72.19 \\
		EPC-Net-L-D (ours) & 87.23 & 80.04 & 75.52 \\
		\hline
		\hline
	\end{tabular}
	\vspace{-15pt}
	\label{tab:results_three}
\end{table}

\begin{figure*}
	\centering
	\includegraphics[width=1.0\textwidth]{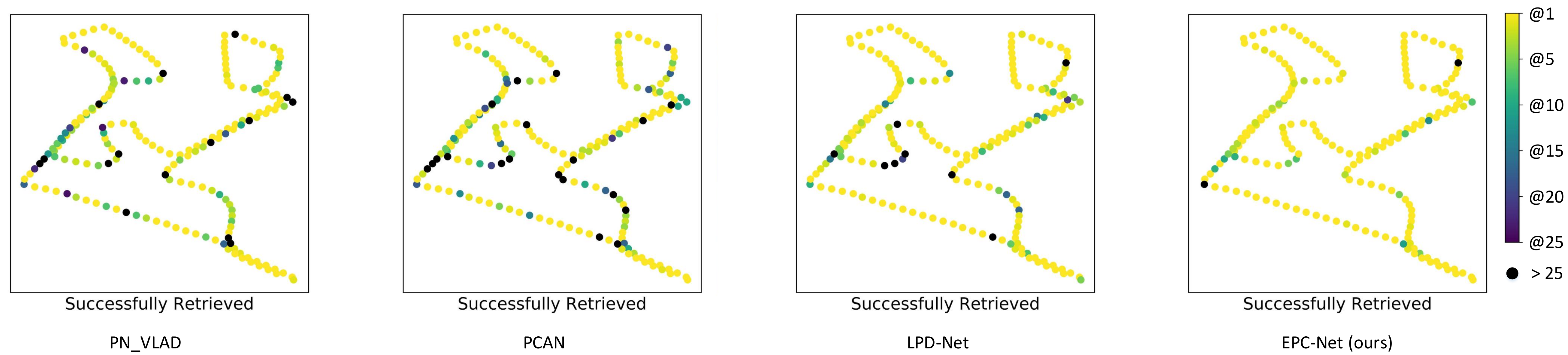}
	\vspace{-20pt}
	\caption{The heat map of different methods for a randomly selected database-query pair of the B.D. dataset. It is shown that our EPC-Net can successfully retrieve almost all the entire scenes.}
	\label{fig:vis_route}
\end{figure*}

\begin{figure*}
	\centering
	\includegraphics[width=0.99\textwidth]{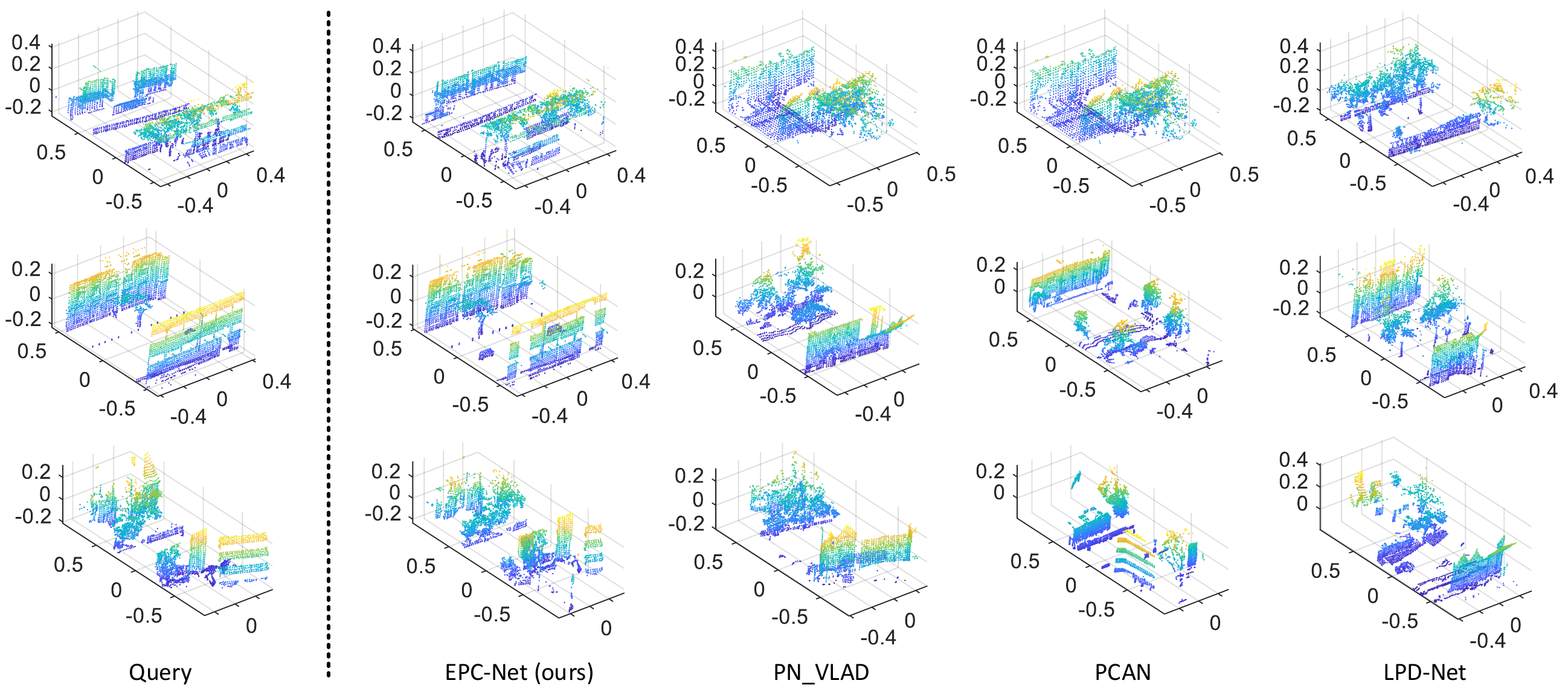}
	\caption{Retrieval results of different methods on Oxford dataset. The leftmost is the query point clouds, and the rest from left to right are the top retrieved point clouds of EPC-Net (ours), PN\_VLAD, PCAN, and LPD-Net, respectively. Note that our method can successfully retrieve the scene of point clouds.}
	\vspace{-10pt}
	\label{fig:vis_retrieval}
\end{figure*}

\subsubsection{Visual results} In Fig.~\ref{fig:vis_route}, we show the heat maps of successfully retrieved submaps of four different methods for a database pair in the B.D. dataset. In the figure, the color close to yellow indicates that the submaps can be recognized more accurately. It can be found that compared to other methods, our EPC-Net has a good ability to identify the places of the entire scenes. As shown in Fig.~\ref{fig:vis_retrieval}, we visualize the retrieval results of different methods for top 1 matching on the Oxford dataset. The visual results reveal that our EPC-Net can successfully retrieve the scenes under different environments compared to other methods. For example, from the first row of the figure, PN\_VLAD and PCAN cannot correctly retrieve the scene, where the environments of the searched scenes are very different from the query scene. Although the scenes searched by LPD-Net also doesn't match the query scene, these scenes are similar in the overall structure. PN\_VLAD and PCAN cannot effectively retrieve the scenes, because they both ignore the local feature learning of the point clouds. LPD-Net uses the dynamic graph in the feature space to extract local features. Obviously, two points close to each other in feature space do not mean that they are close in the spatial space. Note that we use the spatial space to extract the local geometric features in the entire network. Thus, our results are better than LDP-Net. The visual results prove the effectiveness of our method.

\begin{figure*}[t]
	\centering
	\includegraphics[width=1.0\textwidth]{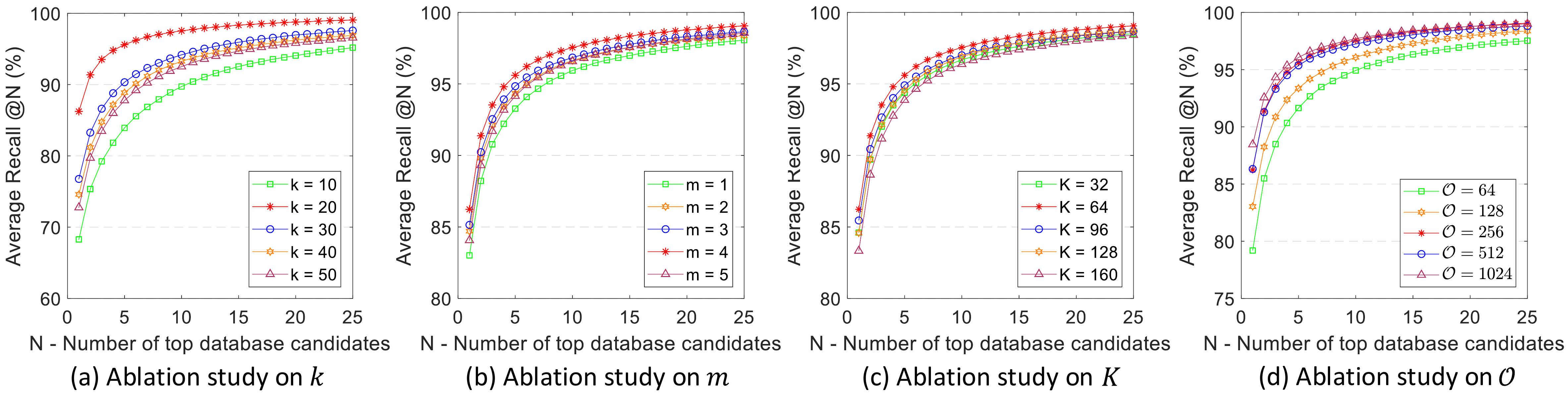}
	\vspace{-20pt}
	\caption{Average recall of different ablation studies on the Oxford dataset. From left to right, they are the results of the number of neighbors $k$, the number of stacking modules, the number of clusters $K$, and the output dimension $\mathcal{O}$, respectively.}
	\vspace{-10pt}
	\label{fig:ablation_curve}
\end{figure*}

\subsubsection{Computational cost} To evaluate the computation and memory required of different methods, we report the number of parameters, floating point operations (FLOPs), and runtime per frame in Tab.~\ref{tab:results_speed}. For a fair comparison, all codes are run on a single NVIDIA TITAN RTX GPU using TensorFlow framework. Thanks to the efficient proxy point convolutional neural network and the G-VLAD network, our EPC-Net can achieve the best result of 94.74\% at a lower number of 4.70M parameters, 3.25G FLOPs, and 32.82ms runtime per frame. Different from PN\_VLAD and LPD-Net, we use the G-VLAD network instead of the original VLAD network. As mentioned before, the number of parameters mainly comes from the last fully connected layer, which maps the high-dimensional vector to the low-dimensional vector. Since the G-VLAD network decomposes the high-dimensional vector into a group of relative low-dimensional vectors by GFC layer, our EPC-Net can obtain a lower number of parameters. In addition, PCAN uses a attention-based VLAD network. Therefore, it has a higher number of parameters than ours. Our EPC-Net still has lower FLOPs and runtime per frame due to the lightweight ProxyConv module. In addition, PN\_VLAD, PCAN and LPD-Net use the spatial transform network (STN) module, which will bring a huge computational cost resulting in higher FLOPs. Note that we do not use STN module in our network. Additionally, LPD-Net uses the EdgeConv modules, so its FLOPs higher than PN\_VLAD. Our FLOPs is lower than other methods due to the ProxyConv module and the grouped fully connected layer. Although the performance of EPC-Net-L is lower than EPC-Net, it still significantly outperforms most of the other methods. Note that in Tab.~\ref{tab:results_speed}, the computation and memory required of EPC-Net-L and EPC-Net-L-D are the same. EPC-Net-L-D has the same structure as EPC-Net-L, but the training is guided by EPC-Net. By distilling the knowledge from EPC-Net, EPC-Net-L-D improves the performance of EPC-Net-L from 86.53\% to 92.23\%, whereas the number of parameters, FLOPs and runtime per frame are the same as EPC-Net-L. Compared with other methods, our EPC-Net-L and EPC-Net-L-D achieve the lowest parameters of 0.41M, 1.37G FLOPs, and 25.71ms runtime per frame, which further proves the effectiveness of our method.

\subsection{Ablation Studies}
\subsubsection{Different $k$ in proxy point convolutional neural network} In our proxy point convolutional neural network, the neighbor size $k$ is an important parameter for extracting local geometric features. Here we study the number of neighbor size $k$ to show the influence in different settings. As shown in Tab.~\ref{tab:ablation_kmKd}, the best results are achieved when the value of $k$ is set to 20. Moreover, we also show the recall curves of different number of $k$ for top 25 retrieval results from the Oxford dataset in Fig.~\ref{fig:ablation_curve}(a). It can be seen that the curve of $k=$ 20 is higher than other curves, which further indicates that it achieves better performance. A smaller $k$ will hurt the performance because it is difficult to generate effective geometric features from a very small neighborhood. However, a larger $k$ will introduce more noise to the neighborhood, thereby hindering the learning of local geometric information. Therefore, we choose $k=$ 20 in the experiment according to Tab.~\ref{tab:ablation_kmKd}. Note that, different from other methods, the choice of $k$ can hardly affect the computational cost in our network. For example, in LPD-Net~\cite{lpdnet2019}, a larger $k$ will generate a larger feature map, which will result in a large amount of computation and memory. However, in our network, only the spatial adjacent matrix contributes to the size of the feature map, rather than the parameter $k$. The size of the adjacent matrix is fixed to $n\times n$ ($n$ is the number of points) so that the different values of $k$ cannot affect the size of the adjacent matrix and the feature map. Thus, the choice of $k$ cannot bring additional computation and memory.

\begin{table}
	\centering
	\caption{Ablation studies of different hyperparameters in our network on the Oxford dataset.}
	\begin{tabular}{l|l|c|c}
		\hline
		\hline
		\multicolumn{2}{c|}{EPC-Net} & Ave Recall @1\% & Ave Recall @1 \\ 
		\hline 
		\multirow{5}{*}{Neighbors} & $k=$ 10 & 81.70 & 68.27 \\ 
		&$k=$ 20 &  {\bf 94.74} & {\bf 86.23} \\
		&$k=$ 30 & 88.68 & 76.76 \\
		&$k=$ 40 & 87.02 & 74.58 \\
		&$k=$ 50 & 85.87 & 72.76 \\
		\hline
		\multirow{5}{*}{Modules} & $m=$ 1 & 92.12 & 83.01 \\ 
		&$m=$ 2 & 93.37 & 84.70 \\ 
		&$m=$ 3 & 93.85 & 85.14 \\ 
		&$m=$ 4 & {\bf 94.74} & {\bf 86.23}  \\ 
		&$m=$ 5 & 93.10 & 84.06 \\ 
		\hline
		\multirow{5}{*}{Output} & $\mathcal{O}=$ 64 & 90.22 & 79.18 \\ 
		&$\mathcal{O}=$ 128 & 92.31 & 83.03 \\ 
		&$\mathcal{O}=$ 256 & 94.74 & 86.23 \\ 
		&$\mathcal{O}=$ 512 & 94.46 &  86.31 \\ 
		&$\mathcal{O}=$ 1024 & {\bf 95.28} & {\bf 88.45} \\ 
		\hline
		\multirow{5}{*}{Clusters} & $K=$ 32 & 93.44 & 84.60 \\ 
		&$K=$ 64 & {\bf 94.74} & {\bf 86.23} \\ 
		&$K=$ 96 & 93.94 & 85.45 \\ 
		&$K=$ 128 & 93.51 & 84.58 \\ 
		&$K=$ 160 & 92.68 & 83.32 \\ 
		\hline
		\hline
	\end{tabular}
	\vspace{-10pt}
	\label{tab:ablation_kmKd}
\end{table}

\subsubsection{Stacking different numbers of ProxyConv modules} In the proxy point convolutional neural network, we stack four ProxyConv modules to obtain multi-scale features to enhance the feature representation. Here, we study the impact of the number of modules used in the network on performance. Specifically, we have tried five schemes using at least one module and up to five modules. For a fair comparison, we only change the number of modules, and the rest of the settings remains the same. Comparison results are show in Tab.~\ref{tab:ablation_kmKd} and Fig.~\ref{fig:ablation_curve}(b), where EPC-Net is equipped with four modules ($i.e.$, $m=$ 4) and achieves the best performance. From the figure, the curve of $m=$ 4 is higher than other curves, which means it has better performance. Since fewer modules will make the receptive field of the network get smaller, we cannot extract robust point features. When stacking more modules, the local receptive field will become larger, which will bring more noise and make it hard to extract effective local geometric information. Therefore, we choose four modules in this paper according to the experiment.

\subsubsection{The effectiveness of proxy point convolutional neural network} In order to compare the feature extraction capabilities of different methods, we conduct experiments on the Oxford dataset to reveal the effectiveness of our PPCNN. Specifically, we use the feature extraction network of PointNetVLAD~\cite{pointnetvlad2018}, PCAN~\cite{pcan2019}, LPD-Net~\cite{lpdnet2019}, and EPC-Net, respectively. For a fair comparison, we use the same VLAD network to obtain global descriptors. In Tab.~\ref{tab:ablation_feature}, we show the average recall and GPU memory of different methods. Note that in the table, we use the method name to represent the corresponding feature extraction network. ``EPC-Net\;+\;VLAD'' means that we use the PPCNN plus the original VLAD network. It can be seen that our ``EPC-Net\;+\;VLAD'' can obtain the best performance with low GPU memory compared to other methods. ``PN\;+\;VLAD'' and ``PCAN\;+\;VLAD'' have similar performance and the same GPU memory, because they both use PointNet as the feature extraction network. Although their GPU memory is 0.5G lower than our method, their performance of the average recall at top 1\% is 12.46\% lower than our method. Since ``LPD-Net\;+\;VLAD'' uses EdgeConv module to extract local geometric features, it has higher performance and GPU memory consumption. Thanks to the ProxyConv module, our method can obtain higher performance compared with ``LPD-Net\;+\;VLAD'' while reducing GPU memory.

\begin{table}
	\centering
	\caption{Ablation studies of different feature extraction network on the Oxford dataset.}
	\resizebox{0.48\textwidth}{!}
	{
		\begin{tabular}{l|c|c|c}
			\hline 
			\hline
			\multicolumn{1}{c|}{Feature Modules} &Ave Recall @1\%&Ave Recall @1& GPU Memory\\ 
			\hline 
			PN\;+\;VLAD & 81.01 & 62.76 & {\bf 4.7G}\\ 
			PCAN\;+\;VLAD & 81.31 & 63.02 & {\bf 4.7G}\\ 
			LPD-Net\;+\;VLAD & 90.88 & 80.71 & 17.8G\\
			DGCNN\;+\;VLAD & 92.56 & 83.04 & 19.5G\\
			\hline
			EPC-Net\;+\;VLAD & {\bf 93.77} & {\bf 85.23} & 5.2G\\ 
			\hline
			\hline
		\end{tabular}
	}
	\vspace{-15pt}
	\label{tab:ablation_feature}
\end{table}

\subsubsection{ProxyConv vs. EdgeConv} In order to compare ProxyConv with EdgeConv, we conduct experiments based on proxy point convolutional neural network (PPCNN) and dynamic graph convolutional neural work (DGCNN~\cite{wang2018dynamic}). Our PPCNN uses ProxyConv modules, whereas DGCNN uses EdgeConv modules. It can be seen from Tab.~\ref{tab:ablation_feature} that our ``EPC-Net\;+\;VLAD'' has comparable results to ``DGCNN\;+\;VLAD'', but has lower GPU memory. On the one hand, because ProxyConv uses the static graph constructed in the spatial space, while EdgeConv uses the dynamic graph constructed in the feature space. Points that are close to each other in the feature space do not mean that they are close to each other in the spatial space. We construct the static graph in the spatial space to better characterize the local geometric structure. On the other hand, there are many planar structures in the scene, as shown in Fig.~\ref{fig:vis_retrieval}. Although EdgeConv can be used on the local planes to extract local features, it suffers from a large amount of GPU memory. The LiDAR points scanned from the local planes are similar to each other. Thus, we adopt the proxy point to replace similar neighbor points in the local plane. In this way, our ProxyConv could learn local features through proxy points while greatly reducing GPU memory. The experimental results further prove that our ProxyConv can achieve comparable results to EdgeConv while having lower GPU memory.



\subsubsection{Different $K$, $\mathcal{O}$, and $G$ in G-VLAD network} In the G-VLAD network, we study the impact of different parameters on network performance. Specifically, we investigate the number of clusters $K$ to explore the impact of the number of local descriptors on aggregating the global descriptor. For a fair comparison, we first fix the number of $G=4$, and $\mathcal{O}=256$, and then select $K\in[32, 64, 96, 128, 160]$. As shown in Tab.~\ref{tab:ablation_kmKd}, we list the results of different number of clusters. The cluster with $K$ = 64 has the best performance. For a more intuitive comparison, we show the recall curves of different number of $K$ for top 25 retrieval results from the Oxford dataset in Fig.~\ref{fig:ablation_curve}(c). It can be seen that the curve of $K=$ 64 higher than other curves. A smaller $K$ will produce rough clustering features, which will lead to poor discrimination during retrieval. However, a larger $K$ will cause the feature fragmentation in the cluster space, which is not conducive to generating robust local descriptors. Thus, according to the figure, we set $K=$ 64 in the experiment. 

Moreover, we also study the discriminative ability of our network on different output dimensions $\mathcal{O}$ of the global descriptor for retrieval. We select the $\mathcal{O}\in[64, 128, 256, 512, 1024]$ in the experiment, and fix the $K=$ 64 and $G=$ 4 for a fair comparison. In Tab.~\ref{tab:ablation_kmKd}, we display the results of different dimensions. Note that the network with output dimension $d=$ 1024 achieves the best recall @1\% and recall @1, which is slightly higher than the network with $\mathcal{O}=$ 256 and $\mathcal{O}=$ 512. In Fig.~\ref{fig:ablation_curve}(d), we also show the curves of different number of $\mathcal{O}$ for top 25 retrieval results from the Oxford dataset. From the figure, the network with $\mathcal{O}=$1024 slightly outperforms the others. However, in the last fully connected layer of the G-VLAD network, the high dimensional output will incur more computation and memory during training. Consequently, for a good trade-off between performance and resource consumption, we set $\mathcal{O}$ to 256 in the experiment.

\begin{table}
	\centering
	\caption{Ablation studies of different number of groups on the Oxford dataset.}
		\begin{tabular}{l|c|c|c}
			\hline
			\hline
			\multicolumn{1}{c|}{\;\;EPC-Net\;\;} & Ave Recall @1\% & Ave Recall @1 & Parameters \\ 
			\hline 
			G = 1 & 93.77 & 85.25 & 17.28M \\ 
			G = 2 & 93.96 & 85.76 & 8.89M \\
			G = 4 & {\bf 94.74} & {\bf 86.23} & 4.70M\\ 
			G = 8 & 93.97 & 85.47 & 2.60M \\ 
			G = 16 & 93.80 & 85.25 & 1.56M \\ 
			G = 32 & 93.52 & 84.98 & {\bf 1.03M} \\ 
			\hline
			\hline
		\end{tabular}
	\vspace{-15pt}
	\label{tab:ablation_G}
\end{table}

To investigate the parameter $G$ in the grouped fully connected (GFC) layer, we conduct experiments under different values of $G$. Specifically, we select the values of $G\in[1, 2, 4, 8, 16, 32]$, and fix the number of $K=$ 64 and $\mathcal{O}=$ 256 meanwhile. The average recall and the parameters on the Oxford dataset are shown in Tab.~\ref{tab:ablation_G}. When $G$ is set to 4, we can achieve the best results compared with other settings, which also means that we can reduce the number of parameters by 4 times. Note that, when $G$ is set to 1, it means that we do not use the group operation to reduce the number of parameters. From the table, it can be found that the result of using the grouping operation is slightly better than the result of not using it ($i.e.$, $G=$ 1), which indicates that our G-VLAD network equipped with the group operation can achieve better performance while having lower parameters. In addition, with the number of groups increase, the number of parameters decreases. When $G=$ 32, the number of parameters can even decrease to 1.03M. Compared with other methods in Tab.~\ref{tab:results_oxford}, we can still achieve the best results, which further demonstrates the effectiveness of our EPC-Net.

\subsubsection{The effectiveness of G-VLAD network} To obtain discriminative global descriptors, we proposed G-VLAD network to replace the original VLAD network for reducing the number of parameters. Here we conduct experiments to study the effectiveness of our G-VLAD network. Specifically, based on our EPC-Net, we use the global average pooling (AVG), the global max pooling (MAX), and the VLAD network (VLAD) to replace the proposed G-VALD network, respectively. As shown in Tab.~\ref{tab:ablation_vlad}, we report the results of different setttings. Note that ``EPC-Net + AVG'' and ``EPC-Net + MAX'' indicate that we simply use the global average pooling and the global max pooling to aggregate the global descriptors. One can see that EPC-Net with G-VLAD achieves the best performance on both average recall at to 1\% and average recall at top 1. More importantly, the number of parameters of our G-VLAD network is much smaller than that of VLAD. Since the grouped fully connected layer decomposes the high-dimensional global vector into a group of low-dimensional vectors, the number of parameters will be reduced by group operations. Although the number of parameter of our G-VLAD is larger than that of the global average pooling and the global max pooling, our performance outperforms them with a large margin. This is because our G-VLAD can extract more discriminative global descriptors than the global average pooling and the global max pooling. Therefore, this further proves the effectiveness of our G-VLAD network.

\begin{table}
	\centering
	\caption{Ablation studies of different descriptors on the Oxford dataset.}
	\resizebox{0.48\textwidth}{!}
	{
		\begin{tabular}{l|c|c|c}
			\hline 
			\hline
			\multicolumn{1}{c|}{Methods} & Ave Recall @1\% & Ave Recall @1 & Parameters \\ 
			\hline 
			EPC-Net\;+\;AVG & 81.61 & 63.81 & {\bf 0.97M} \\ 
			EPC-Net\;+\;MAX & 82.38 & 66.79 & {\bf 0.97M} \\ 
			EPC-Net\;+\;VLAD & 93.77 & 85.23 & 17.28M \\
			\hline
			EPC-Net\;+\;G-VLAD & {\bf 94.74} & {\bf 86.23} & 4.70M\\ 
			
			\hline
			\hline
		\end{tabular}
	}
	\vspace{-10pt}
	\label{tab:ablation_vlad}
\end{table}

\subsubsection{Different $\lambda$ in network distillation} In this paper, we propose a lightweight version of EPC-Net (dubbed ``EPC-Net-L'') to reduce the inference time. We improve the performance of EPC-Net-L (student network) by distilling the knowledge from the EPC-Net (teacher network). Here we study the impact of the parameter $\lambda$ of distilling the knowledge from the teacher network. In the experiment, we chose five different values for $\lambda$ including 0, 0.001, 0.01, 0.1, and 1.0, respectively. As shown in Tab.~\ref{tab:ablation_lambda}, we show the average recall of the network with different settings of the $\lambda$. One can see that when $\lambda$ is set to 0.1, it achieves the best performance. Note that $\lambda=$ 0 means that we only train the student network with the original lazy quadruplet loss. From the table, the student network trained with the guide of the teacher network can improve the performance from 86.53\% to 92.23\%, which further demonstrates the effectiveness of the network distillation strategy.

\section{Conclusion}\label{sec:conclusion}
In this paper, we proposed an efficient point cloud learning network (EPC-Net) for point cloud based place recognition, which can obtain good performance while having few parameters, and short inference time. Specifically, we first proposed a lightweight neural network module, called ProxyConv, which leverages the spatial adjacent matrix and proxy points to simplify the original edge convolution, thereby reducing memory consumption. By stacking the ProxyConv modules, we construct the proxy point convolutional neural network (PPCNN) to aggregate the multi-scale local geometric features while having low resource consumption due to the adjacent matrix shared in all modules. After that, we designed a grouped VLAD network (G-VLAD) to obtain the global descriptors for retrieval. In the G-VLAD network, we used the grouped fully connected (GFC) layers to decompose the high-dimensional vector into a group of relative low-dimensional vectors, thereby reducing the number of parameters. To further reduce the inference time, we also developed a lightweight network (EPC-Net-L), which consists of two ProxyConv modules and one max pooling layer. In order to obtain discriminative global descriptors, we used the EPC-Net to guide the training of EPC-Net-L through knowledge distillation. Extensive experiments on the Oxford dataset and three in-house datasets demonstrated the effectiveness of our proposed method can achieve state-of-the-art performance with lower parameters, FLOPs, and runtime per frame.

\begin{table}[t]
	\centering
	\caption{Ablation studies of different values of $\lambda$ on the Oxford dataset.}
	\begin{tabular}{l|c|c}
		\hline 
		\hline
		\multicolumn{1}{c|}{EPC-Net-L} & Ave Recall @1\% & Ave Recall @1 \\ 
		\hline 
		$\lambda=$ 0 & 86.53 & 73.03 \\ 
		\hline
		$\lambda=$ 0.001 & 87.44 & 73.43 \\ 
		$\lambda=$ 0.01 & 89.40 & 76.51 \\ 
		$\lambda=$ 0.1 & {\bf 92.23} & {\bf 80.34} \\
		$\lambda=$ 1.0 & 82.69 & 66.29 \\ 
		\hline
		\hline
	\end{tabular}
	\label{tab:ablation_lambda}
	\vspace{-10pt}
\end{table}


%





\ifCLASSOPTIONcaptionsoff
  \newpage
\fi



%

\bibliographystyle{IEEEtran}
\bibliography{IEEEtran}

%

\end{document}